# Incremental Model-based Learners
# With Formal Learning-Time Guarantees

**Alexander L. Strehl**
Computer Science Dept.
Rutgers University
Piscataway, NJ 08854 USA
strehl@cs.rutgers.edu

**Lihong Li**
Computer Science Dept.
Rutgers University
Piscataway, NJ 08854 USA
lihong@cs.rutgers.edu

**Michael L. Littman**
Computer Science Dept.
Rutgers University
Piscataway, NJ 08854 USA
mlittman@cs.rutgers.edu

## Abstract

Model-based learning algorithms have been shown to use experience efficiently when learning to solve Markov Decision Processes (MDPs) with finite state and action spaces. However, their high computational cost due to repeatedly solving an internal model inhibits their use in large-scale problems. We propose a method based on real-time dynamic programming (RTDP) to speed up two model-based algorithms, RMAX and MBIE (model-based interval estimation), resulting in computationally much faster algorithms with little loss compared to existing bounds. Specifically, our two new learning algorithms, RTDP-RMAX and RTDP-IE, have considerably smaller computational demands than RMAX and MBIE. We develop a general theoretical framework that allows us to prove that both are efficient learners in a PAC (probably approximately correct) sense. We also present an experimental evaluation of these new algorithms that helps quantify the tradeoff between computational and experience demands.

## 1 Introduction

This paper deals with the important problem of learning how to act in a Markov Decision Process with a finite number of states and actions. This problem is fundamental in reinforcement learning (RL) because it requires the learning agent to handle the *exploration/exploitation dilemma* in conjunction with the *temporal credit assignment* problem. By the exploration/exploitation dilemma, we mean the conflict involved in choosing between a behavior intended to gain new information (explore) and one meant to maximize reward by utilizing current knowledge (exploit). A *sequential decision making* problem involving temporal credit assignment is one where an agent's actions have an effect on the environment and determine which actions and rewards are available in the future.

Agents that learn a model of their unknown environment are called *model-based learners*. We build on previous work in this area and introduce two new model-based learning algorithms, RTDP-RMAX and RTDP-IE. These algorithms distinguish themselves from previous model-based algorithms (such as $R_{\max}$ and $E^3$) in that they avoid completely solving (computing an optimal policy of) their model. This modification alleviates a great computational burden and the resulting algorithms have improved per-action computational complexities. Our main theoretical contribution is showing that the new algorithms still possess polynomial bounds on the amount of mistakes (non $\epsilon$-optimal choices) they make during learning.

## 2 Definitions and Notation

This section introduces the Markov Decision Process notation used throughout the paper; see Sutton and Barto (1998) for an introduction. A finte MDP $M$ is a five tuple $\langle S, A, T, R, \gamma \rangle$, where $S$ is the finite state space, $A$ is the finite action space, $T : S \times A \times S \to \mathbb{R}$ is a transition function, $R : S \times A \to \mathbb{R}$ is a reward function, and $0 \leq \gamma < 1$ is a discount factor on the summed sequence of rewards. We also let $S$ and $A$ denote the number of states and the number of actions, respectively. From state $s$ under action $a$, the agent receives a random reward $r$, which has expectation $R(s, a)$, and is transported to state $s'$ with probability $T(s'|s, a)$. A policy is a strategy for choosing actions. Only deterministic policies are dealt with in this paper. A stationary policy is one that produces an action based on only the current state. We assume (unless noted otherwise) that rewards all lie between 0 and 1. For any policy $\pi$, let $V_M^\pi(s)$ ($Q_M^\pi(s, a)$) denote the discounted, infinite-horizon value (action-value) function for $\pi$ in $M$ (which may be omitted from the notation) from state $s$. If $T$ is a positive integer, let $V_M^\pi(s, T)$ denote the $T$-step value function of policy $\pi$. Specifically, $V_M^\pi(s) = E[\sum_{j=1}^\infty \gamma^{j-1} r_j]$ and $V_M^\pi(s, T) = E[\sum_{j=1}^T \gamma^{j-1} r_j]$ where $[r_1, r_2, \ldots]$ is the reward sequence generated by following policy $\pi$ from state $s$. These expectations are taken over all possible infinite paths the agent might follow. The optimal policy is denoted $\pi^*$ and has value functions $V_M^*(s)$ and $Q_M^*(s, a)$. Note that a policy cannot have a value greater than $1/(1-\gamma)$.

## 3 Problem Definition

In our setting, we assume that the learner receives $S$, $A$, and $\gamma$ as input. The reinforcement-learning problem is defined as follows. The agent always occupies a single state $s$ of the MDP $M$. The learning algorithm is told this state and must compute an action $a$. The agent receives a reward $r$ and is then transported to another state $s'$ according to the reward and transition functions from Section 2. This procedure then repeats forever. The first state occupied by the agent may be chosen arbitrarily. We define a *timestep* to be a single interaction with the environment, as described above.

When evaluating algorithms, there are three essential traits to consider. They are *space complexity*, *computational complexity*, and *sample complexity*. Space complexity measures the amount of memory required to implement the algorithm while computational complexity measures the amount of operations needed to execute the algorithm, per step of experience. Sample complexity measures the number of timesteps for which the algorithm does not behave near optimally or, in other words, the amount of experience it takes to learn to behave well.

We say that an algorithm is *PAC-MDP* (*Probably Approximately Correct in Markov Decision Processes*) if its sample complexity can be bounded by a polynomial in the environment size and approximation parameters, with high probability. Such algorithms, including $E^3$ (Kearns & Singh, 2002), $R_{\max}$ (Brafman & Tennenholtz, 2002), and MBIE (Strehl & Littman, 2005), typically build an internal model of the environment, which is "solved" to find a near-optimal policy.

## 4 Reinforcement-Learning Algorithms

In this section, we present two new RL algorithms that are based on well-known model-based algorithms, $R_{\max}$ and MBIE. The new algorithms, like the old, are based on the idea of using an internal model to guide behavior. However, our main goal is to avoid the computational burden of repeatedly solving a model while preserving the benefits of fast learning by keeping a model. Our main idea can be summed up as applying the Real-time Dynamic Programming (RTDP) algorithm to the agent's internal model. Our approach is similar to that of the Adaptive-RTDP algorithm of Barto et al. (1995), except that more sophisticated exploration techniques are employed.

Our first algorithm, called RTDP-RMAX, uses a model similar to that of the $R_{\max}$ algorithm. The way it treats the exploration problem can be loosely identified as the approach taken by the "naïve" algorithm for the $k$-armed bandit problem (Fong, 1995). Essentially, state-actions must be tried a certain fixed number of times before the statistics gathered for them are incorporated into the agent's model. Until it happens, the state-action is considered to be "maximally rewarding" in the model.

The second algorithm, called RTDP-IE, uses a model similar to that of the MBIE algorithm (Strehl & Littman, 2005) or the Action-Elimination algorithm (Even-Dar et al., 2003). It treats the exploration problem similarly to the Interval Estimation (IE) algorithm for the $k$-armed bandit problem (Kaelbling, 1993). The intuition of the algorithm is that an action is chosen that maximizes the upper tail of a confidence interval computed on the action values maintained by the algorithm. Compared to the naïve algorithm (or $R_{\max}$), this strategy can exhibit more focused behavior and faster learning, as experience is incorporated much faster into the agent's model.

All algorithms we consider maintain *action-value* estimates, $Q(s, a)$ for each state-action pair $(s, a)$. At time $t = 1, 2, \ldots$, let $Q_t(s, a)$ denote the algorithm's current action-value estimate for $(s, a)$ and let $V_t(s)$ denote $\max_{a \in A} Q_t(s, a)$. The learner always acts greedily with respect to its estimates, meaning that if $s_t$ is the $t$th state reached, $a' := \operatorname{argmax}_{a \in A} Q_t(s_t, a)$ is the next action chosen. Additionally, each algorithm makes use of "optimistic initialization", that is, $Q_1(s, a) = 1/(1-\gamma)$ for all $(s, a)$. The main difference between the algorithms is how the action values are updated on each timestep. For all algorithms, we let $n(s, a, t)$ be the number of times $(s, a)$ has been experienced (action $a$ has been taken from state $s$) up to and including timestep $t$ (this corresponds to the first $t$ actions).

### 4.1 The RTDP-RMAX Algorithm

In addition to the standard inputs, the RTDP-RMAX algorithm requires an additional positive integer parameter $m$. Later, in the analysis of Section 5, we provide a formal procedure for choosing $m$, but for now we consider it a free parameter that controls the exploration behavior of the agent (larger values of $m$ encourage greater exploration while smaller values encourage greedier behavior).

Suppose that $a$ is the $t$th action of the agent and is taken from state $s$, and that $n(s, a, t) \geq m$. The following update then occurs:

$$Q_{t+1}(s,a) = \hat{R}_t(s,a) + \gamma \sum_{s' \in S} \hat{T}_t(s'|s,a) V_t(s'), \quad (1)$$

where $\hat{R}_t$ and $\hat{T}_t$ are the empirical reward and transition functions (the maximum likelihood estimates) computed using the history of the agent up to time $t$. For all other $(s', a') \neq (s, a)$, no update occurs. That is, $Q_{t+1}(s', a') = Q_t(s', a')$. Similarly if $n(s, a, t) < m$, then no update occurs.

To summarize, the RTDP-RMAX algorithm chooses, at each step, to either update a single state-action pair or not. If the last state occupied by the agent, under the last action chosen by the agent, has been experienced at least $m$ times, then the action-value estimate for that state-action pair is updated. The update is a standard full Bellman backup, as in value iteration, where the empirical transition probabilities and empirical reward functions are used (in place of the true transition and reward functions, which are unknown by the agent). This approach differs from $R_{\max}$ in that one step of value iteration (VI) is taken from one state instead of running VI to completion.

### 4.2 The RTDP-IE Algorithm

Like RTDP-RMAX, the RTDP-IE (short for "real-time dynamic programming with interval estimation") algorithm also requires an additional real-valued parameter $\beta$ (larger values encourage greater exploration while smaller values encourage greedier behavior) that can be chosen to provide a formal learning-time guarantee, as we show in Section 5.

Suppose that $a$ is the $t$th action by the agent and is taken from state $s$. The following update then occurs:

$$Q_{t+1}(s,a) = \qquad (2)$$
$$\hat{R}_t(s,a) + \gamma \sum_{s' \in S} \hat{T}_t(s'|s,a) V_t(s') + \frac{\beta}{\sqrt{n(s,a,t)}},$$

where $\hat{R}_t$ and $\hat{T}_t$ are the empirical reward and transition functions computed using the history of the agent up to time $t$. For all other $(s', a') \neq (s, a)$, no update occurs. That is, $Q_{t+1}(s', a') = Q_t(s', a')$.

Like RTDP-RMAX, the update is a standard full Bellman backup, where the empirical transition probabilities and reward functions are used, plus an "exploration bonus" proportional to $\beta$ that decreases at a rate inversely proportional to the square root of the number of times the state-action pair has been experienced. Thus, a higher bonus is provided to state-actions that have not been tried as often.

### 4.3 Related Algorithms

For the purpose of a concrete definition, define $R_{\max}$ to be the algorithm that maintains the same model as RTDP-RMAX and that always chooses its actions according to an optimal policy of its model. Similarly, we define MBIE to be the algorithm that maintains the same model[1] as RTDP-IE and acts according to an optimal policy of its model. Specifically, for both $R_{\max}$ and MBIE, the action values, $Q_t(s,a)$, are equal to $Q^*_{\hat{M}_t}(s,a)$, where $\hat{M}_t$ is the agent's model at time $t$. This definition differs slightly from the original development of $R_{\max}$ by Brafman and Tennenholtz (2002) and more significantly from MBIE as presented by Strehl and Littman (2005). However, these differences are not important, in terms of analysis, as the same learning bounds apply to both versions. They are also not likely to be very different experimentally. However, to be completely fair, in our experiments we used the form of the MBIE algorithm as it appears in Strehl and Littman (2005).

The Adaptive-RTDP algorithm is simply the RTDP-RMAX algorithm with $m$ set to 1. The RTDP algorithm can be characterized as being identical to the Adaptive-RTDP algorithm except that the true MDP is used as the model, rather than the empirical model.

### 4.4 Comparison of RTDP-RMAX and RTDP-IE

The two new algorithms we introduce are very similar. They both provide an incentive to explore unknown parts of the environment by increasing the action-value estimate used by the algorithm for state-action pairs that have not been tried that often in the past. This incentive encourages exploration, as the agent chooses actions with maximum current action value, and updates the action values of a state with respect to the action values of the reachable next-states (in proportion to the estimated probability of reaching them). This chain of updates allows the exploration bonuses to propagate from one action value to another and hence encourages directed exploration.

The two algorithms differ in the form of the exploration bonus they provide. The simpler algorithm, RTDP-RMAX, simply enforces an action value of $1/(1-\gamma)$ (the maximum possible true action value) for any state-action pair that has not been tried at least $m$ times. This method works well, but has the drawback of ignoring the statistics collected from the first $m$ tries of each state-action pair. We would expect an intelligent algorithm to take advantage of its experience more quickly. Our second algorithm, RTDP-IE, accomplishes this objective by providing a bonus to each state-action pair that decreases with the number of experiences of that state-action pair. This update allows the experience to be useful immediately, but still recognizes that state-action pairs with little experience need to continue to be explored[2].

---
[1] Here, we add the bonuses, $\beta/\sqrt{n(s,a,t)}$, of Equation 2 to the reward function of the model.

The above intuitions are not new, and are precisely the intuitions used by the $R_{\max}$ algorithm (for RTDP-RMAX) and by the MBIE algorithm (for RTDP-IE). The important difference is that $R_{\max}$ and MBIE work by completely solving their internal model (usually by value iteration, but any technique is valid). This computation incurs a worst-case per-step (action choice) computational cost of $\Omega(SKA)$ (where $K$ is the number of states that can be reached in one step with positive probability), which is highly detrimental in domains with a large number of states and actions. On the other hand, RTDP-RMAX and RTDP-IE require only a single Bellman-style backup, resulting in a per-step computational complexity of $\Theta(K \ln(A))$.[3]

Of great interest is how this tremendous reduction in computational effort affects the performance of the algorithms. This question is very difficult to answer completely. We provide a theoretical analysis and an empirical evaluation that sheds light on the issue. First, we show that it is possible to prove bounds on the number of sub-optimal choices (as formalized in the following section) made by the RTDP-RMAX and RTDP-IE algorithms that are no larger than a constant times the best bounds known for $R_{\max}$ and MBIE, when logarithmic factors are ignored. Second, we evaluate the performance of the four algorithms experimentally on two different MDPs. The empirical results indicate the performance of the new algorithms, in terms of learning, is comparable to the old. Also, as expected, the different algorithms exhibit different tradeoffs between computational and sample complexity.

## 5 Theoretical Analysis

In this section, we provide a detailed analysis of our new algorithms. First, we explain and justify the type of performance bounds we consider. Then, we develop a general framework for proving performance bounds of RL algorithms. This framework allows us to avoid repetition in the analysis of our two algorithms. It also provides methods that can be applied to new RL algorithms as they are discovered in the future. Finally, we apply the techniques to RTDP-RMAX and RTDP-IE, and show that they are efficient learners for MDPs.

To formalize the notion of "efficient learning" we allow the learning algorithm to receive two additional inputs, $\epsilon$ and $\delta$, both positive real numbers. The first parameter, $\epsilon$, controls the quality of behavior we require of the algorithm (how close to optimality do we expect) and the second parameter, $\delta$, which must be less than 1, is a measure of confidence (how certain do we want to be of the algorithm's performance). As these parameters decrease, greater exploration is necessary, as more is expected of the algorithms.

### 5.1 Learning Efficiently

There has been much discussion in the RL community over what defines an efficient learning algorithm and how to define sample complexity. For any fixed $\epsilon$, Kakade (2003) defines the **sample complexity of exploration** (**sample complexity**, for short) of an algorithm $\mathcal{A}$ to be the number of timesteps $t$ such that the non-stationary policy at time $t$, $\mathcal{A}_t$, is not $\epsilon$-optimal from the current state[4], $s_t$ at time $t$ (formally $V^{\mathcal{A}_t}(s_t) < V^*(s_t) - \epsilon$). We believe this definition captures the essence of measuring learning. An algorithm $\mathcal{A}$ is then said to be an **efficient PAC-MDP** (Probably Approximately Correct in Markov Decision Processes) algorithm if, for any $\epsilon$ and $\delta$, the per-step computational complexity and the sample complexity of $\mathcal{A}$ are less than some polynomial in the relevant quantities $(S, A, 1/\epsilon, 1/\delta, 1/(1-\gamma))$, with probability at least $1 - \delta$. It is simply **PAC-MDP** if we relax the definition to have no computational complexity requirement. The terminology, PAC, is borrowed from Valiant (1984), a classic paper dealing with classification.

The above definition penalizes the learner for executing a non-$\epsilon$-optimal policy rather than for a non-optimal policy. Keep in mind that, with only a finite amount of experience, no algorithm can identify the optimal policy with complete confidence. In addition, due to noise, any algorithm may be misled about the underlying dynamics of the system. Thus, a failure probability of at most $\delta$ is allowed. See Kakade (2003) for a full motivation of this performance measure. The analysis of $R_{\max}$ by Kakade (2003) and of MBIE by

---

[2] As our experiments show, it is sometimes the case that the benefit of using RTDP-IE over RTDP-RMAX is small.

[3] The logarithmic dependence on the number of actions is achieved by using a priority queue to access and update the action values.

[4] Note that $\mathcal{A}_t$ is completely defined by $\mathcal{A}$ and the agent's history up to time $t$.

Strehl and Littman (2005) use the same definition as above. The analysis of $R_{\max}$ by Brafman and Tennenholtz (2002) and of $E^3$ by Kearns and Singh (2002) use slightly different definitions of efficient learning.

## 5.2 General Framework

Our theory will be focused on algorithms that maintain a table of action values, $Q(s,a)$, for each state-action pair (denoted $Q_t(s,a)$ at time $t$)[5]. We also assume an algorithm always chooses actions greedily with respect to the action values. This constraint is not really a restriction, since we could define an algorithm's action values as 1 for the action it chooses and 0 for all other actions. However, the general framework is understood and developed more easily under the above assumptions.

**Definition 1** *Suppose an RL algorithm $\mathcal{A}$ maintains a value, denoted $Q(s,a)$, for each state-action pair $(s,a)$ with $s \in S$ and $a \in A$. Let $Q_t(s,a)$ denote the estimate for $(s,a)$ immediately before the tth action of the agent. We say that $\mathcal{A}$ is a greedy algorithm if the tth action of $\mathcal{A}$, $a_t$, is $a_t := \operatorname{argmax}_{a \in A} Q_t(s_t,a)$, where $s_t$ is the tth state reached by the agent.*

The following is a definition of a new MDP that will be useful in our analysis.

**Definition 2** *For an MDP $M = \langle S, A, T, R, \gamma \rangle$, a given set of action values, $Q(s,a)$ for each state-action pair $(s,a)$, and a set $K$ of state-action pairs, we define the known state-action MDP $M_K = \langle S \cup \{s_0\}, A, T_K, R_K, \gamma \rangle$ as follows. Let $s_0$ be an additional state added to the state space of $M$. Under all actions from $s_0$ the agent is returned to $s_0$ with probability 1. The reward for taking any action from $s_0$ is 0. For all $(s,a) \in K$, $R_K(s,a) = R(s,a)$ and $T_K(\cdot|s,a) = T(\cdot|s,a)$. For all $(s,a) \notin K$, $R_K(s,a) = Q(s,a)$ and $T(s_0|s,a) = 1$.*

The known state-action MDP is a generalization of the standard notions of a "known state MDP" of Kearns and Singh (2002) and Kakade (2003). It is an MDP whose dynamics (reward and transition functions) are equal to the true dynamics of $M$ for a subset of the state-action pairs (specifically those in $K$). For all other state-action pairs, the value of taking those state-action pairs in $M_K$ (and following any policy from that point on) is equal to the current action-value estimates $Q(s,a)$. We intuitively view $K$ as a set of state-action pairs for which the agent has sufficiently accurate estimates of their dynamics.

---

[5]The results don't rely on the algorithm having an explicit representation of each action value (for example, they could be implicitly held inside a function approximator).

**Definition 3** *Suppose that for some algorithm there is a set of state-action pairs $K_t$ defined during each timestep $t$. Let $A_K$ be defined as the event, called the escape event, that some state-action pair $(s,a)$ is experienced by the agent at time $t$ such that $(s,a) \notin K_t$.*

Our proofs work by the following scheme (for whatever algorithm we have at hand): (1) Define a set of known state-actions for each timestep $t$. (2) Show that these satisfy the conditions of Proposition 1.

Note that all learning algorithms we consider take $\epsilon$ and $\delta$ as input. We let $\mathcal{A}(\epsilon, \delta)$ denote the version of algorithm $\mathcal{A}$ parameterized with $\epsilon$ and $\delta$. The proof of Proposition 1 follows the structure of the work of Kakade (2003), but generalizes several key steps.

**Proposition 1** *Let $\mathcal{A}(\epsilon, \delta)$ be any greedy learning algorithm such that for every timestep $t$, there exists a set $K_t$ of state-action pairs. We assume that $K_t = K_{t+1}$ unless, during timestep $t$, an update to some action value occurs or the event $A_K$ happens. Let $M_{K_t}$ be the known state-action MDP and $\pi_t$ be the current greedy policy, that is, for all states $s$, $\pi_t(s) = \operatorname{argmax}_a Q_t(s,a)$. Suppose that for any inputs $\epsilon$ and $\delta$, with probability at least $1-\delta$, the following conditions hold for all states $s$, actions $a$, and timesteps $t$: (1) $Q_t(s,a) \geq Q^*(s,a) - \epsilon$ (optimism), (2) $V_t(s) - V^{\pi_t}_{M_{K_t}}(s) \leq \epsilon$ (accuracy), and (3) the total number of updates of action-value estimates plus the number of times the escape event from $K_t$, $A_K$, can occur is bounded by $\zeta(\epsilon, \delta)$ (learning complexity). Then, when $\mathcal{A}(\epsilon, \delta)$ is executed on any MDP $M$, it will follow a $4\epsilon$-optimal policy from its current state on all but*

$$O\left(\frac{\zeta(\epsilon,\delta)}{\epsilon(1-\gamma)^2} \ln \frac{1}{\delta} \ln \frac{1}{\epsilon(1-\gamma)}\right)$$

*timesteps, with probability at least $1 - 2\delta$.*

**Proof sketch:** Suppose $\mathcal{A}(\epsilon, \delta)$ is executed on MDP $M$. Fix the history of the agent up to the $t$th timestep and let $s_t$ be the $t$th state reached. Let $\mathcal{A}_t$ denote the current (non-stationary) policy of the agent. Let $T = \frac{1}{1-\gamma} \ln \frac{1}{\epsilon(1-\gamma)}$. We have that $|V^{\pi}_{M_{K_t}}(s,T) - V^{\pi}_{M_{K_t}}(s)| \leq \epsilon$, for any state $s$ and policy $\pi$ (see Lemma 2 of Kearns and Singh (2002)). Let $W$ denote the event that, after executing policy $\mathcal{A}_t$ from state $s_t$ in $M$ for $T$ timesteps, one of the two following events occur: (a) the algorithm performs a successful update (a change to any of its action values) of some state-action pair $(s,a)$, or (b) some state-action pair $(s,a) \notin K_t$ is experienced

(escape event $A_K$). We have the following:

$$\begin{aligned}
V_M^{\mathcal{A}_t}(s_t, T) &\geq V_{M_{K_t}}^{\pi_t}(s_t, T) - \Pr(W)/(1-\gamma) \\
&\geq V_{M_{K_t}}^{\pi_t}(s_t) - \epsilon - \Pr(W)/(1-\gamma) \\
&\geq V(s_t) - 2\epsilon - \Pr(W)/(1-\gamma) \\
&\geq V^*(s_t) - 3\epsilon - \Pr(W)/(1-\gamma).
\end{aligned}$$

The first step above follows from the fact that following $\mathcal{A}_t$ in MDP $M$ results in behavior identical to that of following $\pi_t$ in $M_{K_t}$ as long as no action-value updates are performed and no state-action pairs $(s, a) \notin K_t$ are experienced. The second step follows from the definition of $T$ above. The third and final steps follow from preconditions (2) and (1), respectively, of the proposition. These hold with probability at least $1 - \delta$.

Now, suppose that $\Pr(W) < \epsilon(1-\gamma)$. Then, we have that the agent's policy on timestep $t$ is $4\epsilon$-optimal: $V_M^{\mathcal{A}_t}(s_t) \geq V_M^{\mathcal{A}_t}(s_t, T) \geq V_M^*(s_t) - 4\epsilon$. Otherwise, we have that $\Pr(W) \geq \epsilon(1-\gamma)$. By an application of the Hoeffding bound, the union bound, and precondition (3) of the proposition, the latter case cannot occur for more than $O(\frac{\zeta(\epsilon,\delta)T}{\epsilon(1-\gamma)} \ln 1/\delta)$ timesteps $t$, with probability at least $1 - \delta$. $\square$

### 5.3 RTDP-RMAX Analysis

For a clean analysis of the RTDP-RMAX algorithm, we need to modify the algorithm slightly. The modification is not necessary when implementing the algorithm (for which we suggest the more natural version as previously described).

Our modification to the original algorithms is as follows. We allow the update, Equation 1, to take place only if the new action value results in a decrease of at least $\epsilon_1$ (we will provide the precise value of $\epsilon_1$ in the proof of Proposition 3). In other words, the following equation must be satisfied for an update to occur:

$$Q_t(s,a) - \left(\hat{R}_t(s,a) + \gamma \sum_{s' \in S} \hat{T}_t(s'|s,a) V_t(s')\right) \geq \epsilon_1. \quad (3)$$

Otherwise, no change is made and $Q_{t+1}(s,a) = Q_t(s,a)$. In addition, the empirical transitions and rewards, $\hat{T}_t$ and $\hat{R}_t$, respectively, are computed using only the first $m$ experiences (next states and immediate rewards) for $(s, a)$. Any additional experiences of $(s, a)$ are discarded and do not affect the model. This second modification has the effect that only a single empirical reward and transition function (per state-action) will be learned by the agent; it does not continue to adjust its model over time. Hence, we will use the simpler notation $\hat{T}$ and $\hat{R}$. We also let $\hat{M}$ denote the empirical MDP.

We expect that the modifications above will typically have little effect on the behavior and performance (as measured by discounted reward or sample complexity) of the algorithm, but, since they bound the number of times the model can change during learning, simplify the analysis. The condition on the update will only affect updates that would have resulted in a minimal change to the action-value estimates. For sufficiently large $m$, further refinements to the model also have only a minor effect.

During timestep $t$ of the execution of RTDP-RMAX, we define $K_t$ to be the set of all state-action pairs $(s, a)$, with $n(s, a, t) \geq m$ such that:

$$Q_t(s,a) - \left(\hat{R}(s,a) + \gamma \sum_{s'} \hat{T}(s'|s,a) V_t(s')\right) \leq \epsilon_1. \quad (4)$$

Note that $\hat{M}_{K_t}$ is a well-defined known-state MDP with respect to the agent's model at time $t$ (see Definition 2). It can be viewed as an approximation to $M_{K_t}$. This novel definition extends the standard definition (as used in the analysis of $R_{\max}$ and MBIE), which associates $K_t$ with the state-action pairs that have been tried $m$ times, to allow incremental updates to propagate value information more gradually.

The following condition will be needed for our proof that RTDP-RMAX is PAC-MDP. We will provide a sufficient condition (specifically, $L_1$-accurate transition and reward functions) to guarantee that it holds. In words, the first part of the condition says that the value of the greedy policy (with respect to the agent's action values) in the empirical known state-action MDP ($\hat{M}_{K_t}$) is $\epsilon_1$-close to its value in the true known state-action MDP ($M_{K_t}$). The second part says that the optimal value function of the last and final model learned by RTDP-RMAX is not too far from what it would be if the correct transitions and rewards were used for those state-actions tried at least $m$ times.

**Assumption A1** *For all timesteps $t$ and states $s$, we have that $|V_{M_{K_t}}^{\pi_t}(s) - V_{\hat{M}_{K_t}}^{\pi_t}(s)| \leq \epsilon_1$ where $\pi_t$ is the greedy policy (with respect to the agent's action-value estimates) at time $t$. Also, $|V_{M_{\tilde{K}}}^*(s) - V_{\hat{M}_{\tilde{K}}}^*(s)| \leq \epsilon_1$ where $\tilde{K} = \{(s, a) \mid \exists u \in \mathbb{Z}^+ \text{ s.t. } n(s, a, u) \geq m\}$.*

We are now ready to show that the RTDP-RMAX algorithm exhibits the property of "optimism".

**Proposition 2** *Suppose RTDP-RMAX is executed on any MDP $M$. If Assumption A1 holds, then $Q_t(s,a) \geq Q^*(s,a) - \epsilon_1$ holds for all timesteps $t$ and state-action pairs $(s, a)$.*

**Proof:** Let $M'$ denote the *final* model of the algo-rithm (such a model exists because RTDP-RMAX can update its model at most $SA$ times). Note that the update, Equation 1, is identical to the update used by value iteration (called the Bellman update) on the MDP $M'$. It is well known, see Chapter 2 of Bertsekas and Tsitsiklis (1996), that when given optimistic initialization ($Q_1(s, a) = 1/(1 − \gamma)$), any sequence of such updates cannot drive the action values below the optimal $Q^*$-values of $M'$. By Assumption A1, the optimal $Q^*$-values of $M'$ are no less than the the optimal $Q^*$-values of the true MDP $M$ minus $\epsilon_1$, which yields the result. □

**Proposition 3** *If, on execution of the RTDP-RMAX algorithm in any MDP $M$, Assumption A1 holds with probability at least $1 − \delta/2$, then the RTDP-RMAX algorithm is efficient PAC-MDP.*

**Proof:** We apply Proposition 1. First, note that by definition the algorithm performs updates if and only if the event $A_K$ occurs and the current state-action pair has been tried at least $m$ times. Hence, it is sufficient to bound the number of times that $A_K$ occurs. Once a state-action pair has been tried $m$ times, every additional update decreases its action value by at least $\epsilon_1$. Since its action value is initialized to $1/(1 − \gamma)$, we have that each state-action pair can be updated at most $1/(\epsilon_1(1 − \gamma))$ times, for a total of at most $SAm + SA/(\epsilon_1(1 − \gamma))$ timesteps $t$ such that $A_K$ can occur. By Proposition 2, we have that the optimism precondition is satisfied. Finally, we claim that, by Assumption A1, $V_t(s) − V^{\pi_t}_{M_{K_t}}(s) \le 2\epsilon_1/(1 − \gamma)$ always holds. To verify this claim, note that $V^{\pi_t}_{\hat{M}_{K_t}}$ is the solution to the following set of equations:

$$V^{\pi_t}_{\hat{M}_{K_t}}(s) = \hat{R}(s, \pi_t(s)) + \gamma \sum_{s' \in S} \hat{T}(s'|s, \pi_t(s)) V^{\pi_t}_{\hat{M}_{K_t}}(s'),$$

$$\text{if } (s, \pi_t(s)) \in K_t,$$

$$V^{\pi_t}_{\hat{M}_{K_t}}(s) = Q_t(s, \pi_t(s)), \quad \text{if } (s, \pi_t(s)) \notin K_t.$$

The vector $V_t$ is the solution to a similar set of equations except with some additional positive reward terms, each bounded by $\epsilon_1$ (see Equation 4). It follows that $V_t(s) − V^{\pi_t}_{\hat{M}_{K_t}}(s) \le \epsilon_1/(1 − \gamma)$. Combining this fact with Assumption A1 yields $V_t(s) − V^{\pi_t}_{M_{K_t}}(s) \le 2\epsilon_1/(1−\gamma)$. Thus, by letting $\epsilon_1 = \epsilon(1−\gamma)/2$, we satisfy $V_t(s) − V^{\pi_t}_{M_{K_t}}(s) \le \epsilon$, as desired (to fulfill Condition (2) of Proposition 1). Ignoring log factors, this analysis leads to a total sample complexity bound of

$$\tilde{O}\left(\left(SAm + \frac{SA}{\epsilon(1 − \gamma)^2}\right) \frac{1}{\epsilon(1 − \gamma)^2}\right). \quad (5)$$

□

### 5.3.1 Sufficient Condition for Assumption A1 to Hold

One way to guarantee that, with probability at least $1 − \delta/2$, Assumption A1 will hold on execution of RTDP-RMAX is by requiring that

$$m = \Theta\left(\frac{S}{\epsilon^2(1 − \gamma)^4} \ln\left(\frac{SA}{\delta}\right)\right) = \tilde{O}\left(\frac{S}{\epsilon^2(1 − \gamma)^4}\right).$$

With this many samples, the empirical model $\hat{R}(s, a)$ and $\hat{T}(\cdot|s, a)$ will be close enough (in $L_1$ distance for the transitions) to the true dynamics $R(s, a)$ and $T(\cdot|s, a)$ (Kakade, 2003; Strehl & Littman, 2005). The argument boils down to showing that $\Theta(\epsilon(1 − \gamma)^2)$-accurate rewards and transitions lead to $\epsilon$-optimal policies, and that $\Theta(S \ln(1/\delta)/\alpha^2)$ samples are needed to ensure $\alpha$-accurate rewards and transitions, with high probability. The additional log factors are due to an application of the union bound for the result to hold over *all* state-action pairs.

### 5.3.2 Remarks on Our Speedup

By applying Proposition 3 with the value of $m$ from Section 5.3.1, we have shown that RTDP-RMAX is PAC-MDP with a sample complexity bound of

$$\tilde{O}\left(\frac{S^2 A}{\epsilon^3(1 − \gamma)^6}\right). \quad (6)$$

This expression matches the best known bound for $R_{\max}$ (Kakade, 2003) and MBIE (Strehl & Littman, 2005), when logarithmic factors are ignored. Although we don't have room for details, we can prove similar bounds for RTDP-IE.

It is an open question whether the current bounds for $R_{\max}$ can be improved in terms of its dependence on $S$ (Kakade, 2003). In his thesis, Kakade provides a sufficient condition (Condition 8.5.1) that the parameter $m$ must satisfy for his PAC-MDP proof of $R_{\max}$ to go through. Our condition, Assumption A1, is stronger, but similar in spirit to Condition 8.5.1.

### 5.4 RTDP-IE Analysis

The analysis of RTDP-IE actually follows that of RTDP-RMAX very closely. As in Section 5.3, we modify the algorithm slightly.

In the version of RTDP-IE that we analyze, an update is performed as specified by Equation 2 only if it would result in a decrease of at least $\epsilon_1$. In addition, the empirical transitions and rewards, $\hat{T}_t$ and $\hat{R}_t$, respectively, are computed using only the first $m$ experiences ($m$ is an additional parameter supplied to the algorithm) for $(s, a)$. Furthermore, once $(s, a)$ has been experienced $m$ times, the bonus of $\beta/\sqrt{n(s,a,t)}$ in Equation 2 is replaced by $\epsilon_1$. The full proofs of this section can be found in our extended paper.

**Proposition 4** *For any $0 < \delta < 1$ and positive integer $m$, if Assumption A1 holds and $\beta \geq (1/(1-\gamma))\sqrt{\ln(SAm/\delta)/2}$, then during execution of RTDP-IE, with probability at least $1-\delta/2$, $Q_t(s,a) \geq Q^*(s,a)$ holds for all state-action pairs $(s,a)$ and timesteps $t$.*

**Proposition 5** *There exist values for $m$ and $\beta$ such that the RTDP-IE algorithm is efficient PAC-MDP with a sample complexity bound of $\tilde{O}\left(\frac{S^2 A}{\epsilon^3(1-\gamma)^6}\right)$.*

## 6 Experiments

To better quantify the relationship between them, we performed two sets of experiments, each with two versions of all four algorithms ($R_{\max}$, MBIE, RTDP-RMAX, and RTDP-IE). In the first version, we severely restricted the model size (the maximum number of next-state and immediate reward samples per state-action pair that are used for computing the empirical transition and reward functions). Specifically, the model was limited to size $m$ for $R_{\max}$ and RTDP-RMAX and to size 3 for MBIE and RTDP-IE, per state-action pair. In the second version, we allowed the model to grow[6] up to size 100 (per state-action). There numbers were chosen somewhat arbitrarily to be "small" and "big". We experimented over a range of the various parameter settings ($m$ for $R_{\max}$ and RTDP-RMAX and $\beta$ for MBIE and RTDP-IE) and optimized for the parameter setting that gathered the most reward in the fewest timesteps.

For each experiment, we recorded the number of timesteps (to measure sample complexity) and Bellman backups (to measure computational complexity) required by the agent to achieve fixed and equally spaced levels of cumulative reward. We implemented various optimizations on each algorithm to avoid unnecessary Bellman backup computations. Our experiments were designed to supplement the theoretical analysis of Section 5 and are not meant as a thorough evaluation of our new algorithms.

The algorithms were first tested on random MDPs generated as follows, for parameters $S = 50$, $A = 5$, $\gamma = 0.95$, and $R_{\max} = 1$. To guarantee that every state was reachable from every other state, random Hamiltonian circuits (to ensure connectivity) were constructed under each action, and a probability of 0.1 was assigned to each of the corresponding transitions. Then, for each state-action pair, the transition function was completed by assigning random probabilities (summing up to 0.9) to 4 randomly selected next states. Therefore, for each state-action pair, there were at most 5 next-states. The mean reward, $R(s,a)$, was randomly selected with mean proportional to the index of $s$.[7] Each experiment was repeated 100 times and the results were averaged (error bars were omitted from the figure because they were tiny).

[6]Allowing the model size to change independently of the algorithm's internal parameters required a slight change to the algorithm in some cases. For example, for $R_{\max}$, the agent's model may change for a state-action pair that is already "known". In this case, the agent must solve the model each time it changes. Thus, allowing the model to grow may increase the computational complexity, but will also generally decrease the sample complexity.

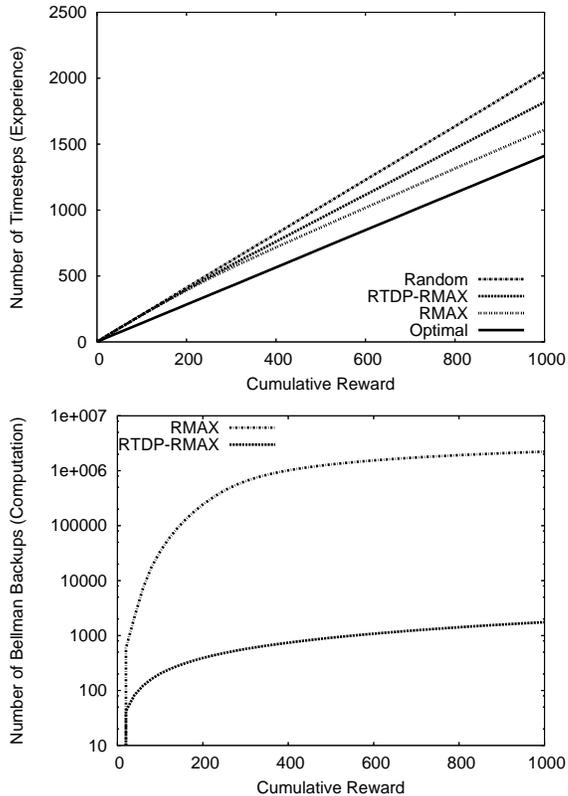

Figure 1: Algorithmic performance on Random MDPs

Figure 1 provides a plot of the runs for the algorithms whose models were restricted to 100 samples (the results with more restricted models were similar). Since the curves for the $R_{\max}$ and MBIE algorithms were so alike we omitted the one for MBIE (similarly for RTDP-RMAX and RTDP-IE). Each graph contains one data point for every 20 units of cumulative reward (per algorithm). From the top graph, we see that MBIE/$R_{\max}$ was able to obtain more cumulative reward per timestep (on average) than the incremental algorithms, but the bottom graph shows that this achievement came at significantly larger computational cost.

[7]We found that if the mean rewards are chosen uniformly at random, then the optimal policy almost always picks the action that maximizes the immediate reward, which is not interesting in the sense of sequential decision making.

The second set of experiments consisted of an MDP similar to the $k$-armed bandit problem (with $k = 6$), where the noise of the arms was modeled in the transition function (rather than the reward function). Specifically, there were 7 states ($S = \{0, \ldots, 6\}$), with 0 as the start state and 6 actions. Taking action $j \in \{1, \ldots, 6\}$ from state 0 results in a transition to state $j$ with probability $1/j$ and a transition back to state 0 with probability $1 - 1/j$. From state $i > 0$, under each action, the agent is transitioned to state 0. Choosing action 1 from any state $i > 0$ results in a reward of $(3/2)^i$ (all other rewards are 0). These dynamics were created so that it is better to choose the action with the lowest payoff probability (leaving state 0). To recap, from state 0, each action behaves like pulling a "one-armed bandit". The arm "pays off" if the agent is transitioned away from state 0. Once in another state, the agent is free to choose action 1 and obtain non-zero reward. Each experiment was repeated 500 times and the results averaged.

The following results, for the algorithms with severely restricted models (as described above), provide the number of timesteps and backups required by each algorithm (at its optimal parameter setting) to obtain a total reward of 15000:

|            | Param | Timesteps | Backups |
|------------|-------|-----------|---------|
| Optimal    | -     | 9213      | 0       |
| $R_{\max}$ | 6     | 12129     | 8761    |
| MBIE       | 0.7   | 12914     | 4406    |
| RTDP-IE    | 0.9   | 13075     | 5558    |
| RTDP-RMAX  | 4     | 13118     | 5618    |
| Random     | -     | 90252     | 0       |

When allowed to learn a less restricted model of size 100 (to avoid repetition, we omitted the optimal and random agents) the following results were obtained:

|            | Param | Timesteps | Backups |
|------------|-------|-----------|---------|
| MBIE       | 0.05  | 10135     | 603513  |
| RTDP-IE    | 0.2   | 11042     | 4391    |
| RTDP-RMAX  | 1     | 11127     | 4438    |
| $R_{\max}$ | 9     | 11286     | 336384  |

We see that all algorithms improved their performance by using a more refined model, but RTDP-IE and RTDP-RMAX could do so without as large an increase in computation[8]. The differences in the number of timesteps used by the RTDP-IE, RTDP-RMAX, and $R_{\max}$ algorithms, shown in the previous table, were not significant.

## 7 Conclusion

We have shown that provably efficient model-based reinforcement learning can be achieved without the computational burden of completely solving the internal model at each step. To do so, we developed two new RL algorithms, RTDP-RMAX and RTDP-IE, and analyzed their sample complexity in general MDPs.

## Acknowledgments

Thanks to the National Science Foundation (IIS-0325281). We also thank John Langford and Eric Wiewiora for discussions.

## References

Barto, A. G., Bradtke, S. J., & Singh, S. P. (1995). Learning to act using real-time dynamic programming. *Artificial Intelligence*, *72*, 81–138.

Bertsekas, D. P., & Tsitsiklis, J. N. (1996). *Neuro-dynamic programming*. Belmont, MA: Athena Scientific.

Brafman, R. I., & Tennenholtz, M. (2002). R-MAX—a general polynomial time algorithm for near-optimal reinforcement learning. *Journal of Machine Learning Research*, *3*, 213–231.

Even-Dar, E., Mannor, S., & Mansour, Y. (2003). Action elimination and stopping conditions for reinforcement learning. *The Twentieth International Conference on Machine Learning (ICML 2003)* (pp. 162–169).

Fong, P. W. L. (1995). A quantitative study of hypothesis selection. *Proceedings of the Twelfth International Conference on Machine Learning (ICML-95)* (pp. 226–234).

Kaelbling, L. P. (1993). *Learning in embedded systems*. Cambridge, MA: The MIT Press.

Kakade, S. M. (2003). *On the sample complexity of reinforcement learning*. Doctoral dissertation, Gatsby Computational Neuroscience Unit, University College London.

Kearns, M. J., & Singh, S. P. (2002). Near-optimal reinforcement learning in polynomial time. *Machine Learning*, *49*, 209–232.

Strehl, A. L., & Littman, M. L. (2005). A theoretical analysis of model-based interval estimation. *Proceedings of the Twenty-second International Conference on Machine Learning (ICML-05)* (pp. 857–864).

Sutton, R. S., & Barto, A. G. (1998). *Reinforcement learning: An introduction*. The MIT Press.

Valiant, L. G. (1984). A theory of the learnable. *Communications of the ACM*, *27*, 1134–1142.

---

[8]We do not totally understand why the total computation of RTDP-IE and RTDP-RMAX went down when the model was increased. It is certainly not true in general.